\documentclass[9pt,twocolumn,letter]{article}

\usepackage{wacv}      

\usepackage{graphicx}
\usepackage{amsmath}
\usepackage{amssymb}
\usepackage{booktabs}
\usepackage{algorithm}
\usepackage{svg}
\usepackage{algpseudocode}
\usepackage{multirow}
\usepackage{array}
\usepackage{setspace}
\usepackage{tikz}
\usepackage{titlesec}
\usepackage{caption}
\usepackage{fancyhdr}

\titlespacing*{\section}{0pt}{0.5em}{0.5em}
\titlespacing*{\subsection}{0pt}{0.5em}{0.5em}

\usetikzlibrary{shapes, arrows, positioning}

\usetikzlibrary{3d,shapes,arrows.meta,positioning,backgrounds}
\graphicspath{ {./images/} }

\usepackage[pagebackref,breaklinks,colorlinks]{hyperref}

\usepackage[capitalize]{cleveref}
\crefname{section}{Sec.}{Secs.}
\Crefname{section}{Section}{Sections}
\Crefname{table}{Table}{Tables}

\begin{document}
\topmargin=0mm
\title{ATAC-Net: Zoomed view works better for Anomaly Detection}

  
\name{Shaurya Gupta, Neil Gautam\sthanks{Work done during internship with HyperVerge}, Anurag Malyala}

\address{HyperVerge Inc}


\maketitle
\begin{abstract}
   The application of deep learning in visual anomaly detection has gained widespread popularity due to its potential use in quality control and manufacturing. Current standard methods are Unsupervised, where a clean dataset is utilised to detect deviations and flag anomalies during testing. However, incorporating a few samples when the type of anomalies is known beforehand can significantly enhance performance. Thus, we propose ATAC-Net, a framework that trains to detect anomalies from a minimal set of known prior anomalies. Furthermore, we introduce attention-guided cropping, which provides a closer view of suspect regions during the training phase. Our framework is a reliable and easy-to-understand system for detecting anomalies, and we substantiate its superiority to some of the current state-of-the-art techniques in a comparable setting.
\end{abstract}

\begin{keywords}
 anomaly detection, self-explainability, weak supervision, attention augmentation, deviation loss
\end{keywords}

\footnote{© 20XX IEEE. Personal use of this material is permitted. Permission from IEEE must be obtained for all other uses, in any current or future media, including reprinting/republishing this material for advertising or promotional purposes, creating new collective works, for resale or redistribution to servers or lists, or reuse of any copyrighted component of this work in other works.}

\section{Introduction}
\label{sec:intro}
Deep learning has achieved state-of-the-art results, matching or surpassing human-level performance in various areas such as healthcare, autonomous driving, natural language processing, and computer vision tasks. Even though these deep learning networks perform better when increasing the available size of the datum, only some domains have enough data for every class in the dataset, leading to significant imbalances in the data distributions and thus biases. This paper thoroughly examines anomaly detection as one of these domains.

\begin{figure}
    \centering
    \includegraphics[width=0.71\linewidth]{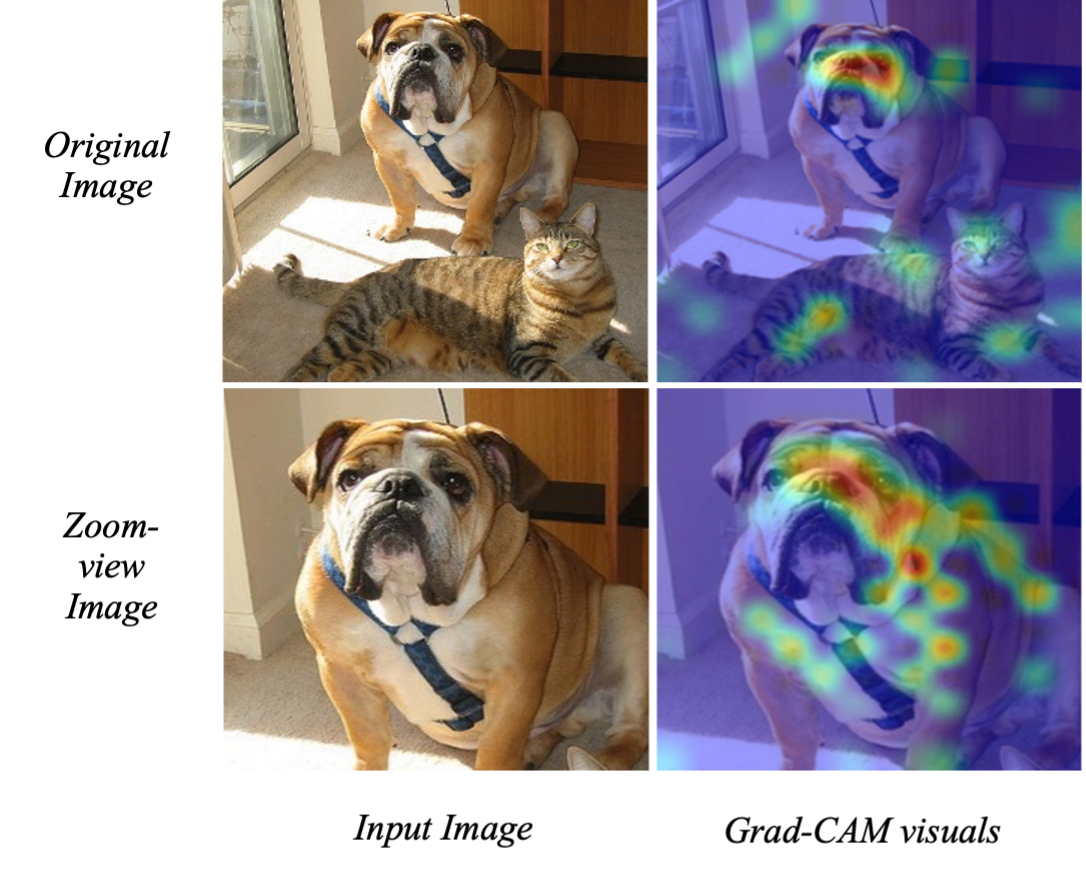} 
    \caption{Grad-CAM \cite{grad-cam} visualization of a dog class within the original image and zoomed view of the object. The density of the saliency map increases as the model receives a zoomed object  }
    \label{fig:fig-1-grad-cam}
\end{figure}

Anomaly detection involves identifying deviations from the typical sample distribution through an anomaly score's measure. Despite its potential, the area of anomaly detection utilising deep learning techniques has yet to be extensively researched, mainly due to specific key characteristics of the data.  


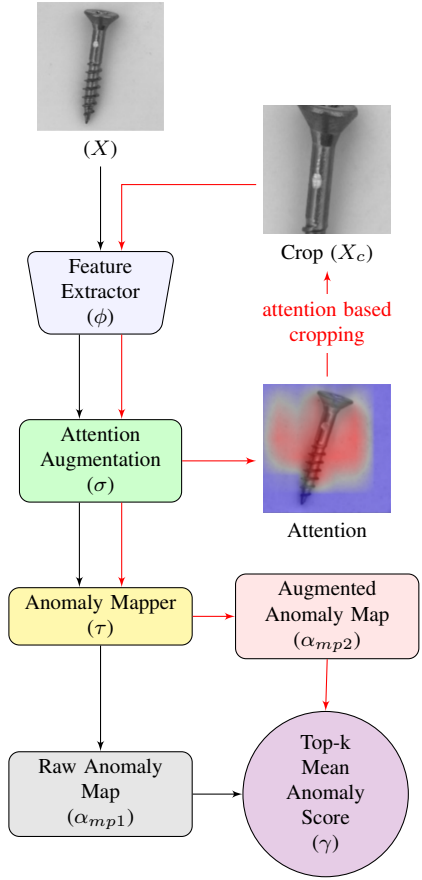
\begin{figure}[h]
    \centering
    \footnotesize
    \begin{tikzpicture}
        \tikzstyle{input-node} = [rectangle,  text centered, text width=6em]
        \tikzstyle{cnn} = [trapezium, trapezium angle=80, draw, shape border rotate=180, fill=blue!05, rounded corners, text width=5em, text centered]
        \tikzstyle{stage} = [rectangle, text centered, text width=7em, rounded corners, draw]
        \tikzstyle{loss} = [circle, text centered, text width=4em, draw]

        \node[input-node](input-image){\includegraphics[width=6em]{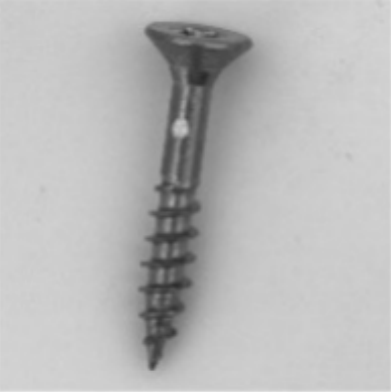}  \\  ($X$)};

        \node[cnn, below=4em of input-image](feature){Feature Extractor ($\phi$)};

        \node[stage, fill=green!20, below = 4em of feature](attention){Attention Augmentation \\ ($\sigma$)};

        \node[input-node, right=3.5em of attention](attended){\includegraphics[width=6em]{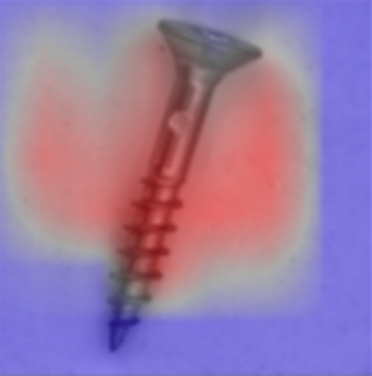} \\ Attention};

        \node[input-node, above=5em of attended](crop){\includegraphics[width=6em]{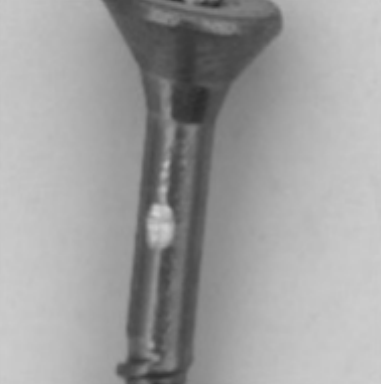} \\ Crop ($X_c$)};

        \node[stage, below=4em of attention, text width=8em, fill=yellow!40](mapper){Anomaly Mapper \\ ($\tau$)};
        \node[stage, right=2em of mapper, text width=8em, fill=red!10](alpha2){Augmented Anomaly Map \\ ($\alpha_{mp2}$)};
        \node[stage, below=5em of mapper, text width=8em, fill=black!10](alpha1){Raw Anomaly Map \\ ($\alpha_{mp1}$)};

        \node[loss, right=2.35em of alpha1, fill=violet!20](loss-val){Top-k Mean Anomaly Score \\ ($\gamma$)};

        \path[draw, -latex'] (input-image.south) -- (feature.north);
        
        \path (feature.north) ++(1em,0em) coordinate (end);
        \path[draw, -latex', red] (crop.west) -| (end);
        
        \path[draw, -latex', transform canvas={xshift=-1em}] (feature.south) -- (attention.north);
        \path[draw, -latex', transform canvas={xshift=1em}, red] (feature.south) -- (attention.north);        
        \path[draw, -latex', red] (attended.north) -- (crop.south) node[midway, fill=white, text width=8em, text centered] {attention based cropping};
        \path[draw, -latex', red] (attention.east) -- (attended.west);

        \path[draw, -latex', transform canvas={xshift=1em}, red] (attention.south) -- (mapper.north);
        \path[draw, -latex', transform canvas={xshift=-1em}] (attention.south) -- (mapper.north);

        \path[draw, -latex', red] (mapper.east) -- (alpha2.west);
        \path[draw, -latex'] (mapper.south) -- (alpha1.north);
        
        \path[draw, -latex'] (alpha1.east) -- (loss-val.west);
        \path[draw, -latex', red] (alpha2.south) -- (loss-val.north);
        
    \end{tikzpicture}

    \caption{ATAC-Net, using attention augmentation module to learn and find anomalies while also providing interpretability of the anomaly via the saliency map. The network’s feature extraction works two-fold to find the best results for anomalies within the provided sample. The normal (un-cropped) and attention-cropped flows are used in train and inference time.}
    \label{fig:fig-2-atac-net}
\end{figure}

Anomaly detection datasets often have only two classes, "normal" and "anomalous" samples, with the number of "normal" samples significantly exceeding that of "anomalous" samples by a factor of at least 100. Moreover, anomalies can greatly differ from one another or exhibit vast dissimilarities. Coupled with a limited number of anomaly samples, it creates a complex problem.

Various methods have been developed to address the problem at hand without the need for supervision. Recent techniques such as \cite{total_recall} and \cite{fflow} fall under unsupervised approaches, which do not require anomalous samples during training. Unsupervised approaches utilise feature extraction and deviation calculation based on anomaly scores, reconstruction errors, or distance-based scores. However, the most significant challenge is determining the optimal threshold, as genuine samples with noise can sometimes get misclassified as anomalies.

This work delves into the domain of anomaly detection using weak supervision. We suggest attention-guided cropping as an augmentation method to enable the model to concentrate better on object features. Our findings, as illustrated in \cref{fig:fig-1-grad-cam}, demonstrate that the model can concentrate more on the object's features when presented with a focused or cropped image rather than the raw input image. The model performed better with zoomed inputs, particularly the object's Region of Interest (ROI). By guiding the model in this manner, we achieved a higher accuracy in anomaly detection, meeting the benchmark accuracy on three distinct datasets. In this paper, we contribute the following insights:

\begin{itemize}
    \setlength\itemsep{-0.5em}
    \item We propose a weakly supervised attention augmentation method to guide the learning of the anomaly score network by focusing on regions of interest (ROI) in the input image.
    \item We use deviation-loss modelling, based on the Gaussian distribution, to structure the anomaly scores accurately.
    \item We tested on three public real-world datasets to test the generalisation ability of the proposed approach and achieved benchmark results when compared to state-of-the-art methods.
\end{itemize}

\section{Literature Review}
\label{sec:lit-review}

In this section, we delve into the evolution of anomaly detection research, focusing on supervised and unsupervised learning methods. Early works in anomaly detection focused on density-based and distance-based modelling techniques. However, these methods proved inadequate in handling high resolutions and failed to effectively map non-linear relationships among the data.

PaDiM, introduced by Defard et al. \cite{padim}, utilises an unsupervised domain to compare query image patches with a feature database. The training process involves representing each image patch using a multivariate Gaussian Distribution with intermediate layer features. During testing, PaDiM compares the features of the query image with the learned distribution using Mahalanobis Distance \cite{mahab-dist}. The PaDiM model has several variants, each with a different backbone type. This paper compares the ResNet-50 \cite{resnet} variant against other options.

PatchCore, by Roth et al. \cite{total_recall}, involves using a memory bank of normal samples to compare with a query sample during testing. The memory saves the patch-aware features, which get downsized using greedy corset sampling. Each query patch is compared during query time to determine if it is an anomaly. Another approach, proposed by Zhang et al. \cite{memseg}, called MemSeg, which combines the memory bank features of normal samples with reconstruction methods. They introduce anomalies only in external datasets in the foreground of items using various techniques to generate noise over genuine samples. Another technique called FastFlow, introduced by Jiawei et al. \cite{fflow}, uses normalising flows \cite{norm-flows} to transform the input to a normal distribution, forcing the anomalies to have mappings distant from the genuine samples, making detection easier. Before this, Gudovskiy et al. \cite{cflowad} proposed a method called CFlow-AD, which uses Condition Normalising Flows. CFlow-AD learns the probability distribution of the genuine samples fed into the model to convert the original distribution to a Gaussian density. This method distinguishes between out-of-distribution patches from normal ones.

Ding et al. \cite{ding-shen} have achieved state-of-the-art results in supervised learning-based approaches. Their novelty lies in using four learning heads, each with a different objective, to optimise the model and improve convergence. Additionally, they utilised a few anomalous samples for training and employed the deviation loss function proposed by Pang et al. \cite{pang-devloss} for anomaly detection through supervised training. Our proposed technique also uses this loss function.

Other approaches, such as those based on Focal \cite{focal-loss} and L1 loss \cite{zhao-l1-loss}. For instance, Zhang et al. \cite{Zhang2022PrototypicalRN} introduced Prototypical Res-Nets (PRN) that learn multi-scale features with Focal and L1 loss during training with labelled segment maps as ground truth. To address data imbalance, they created pseudo anomalies.


\section{Methodology}
\label{sec:method}

This section provides the proposed technique and intuition behind using attention cropping with deviation loss \cite{pang-devloss} for weak supervision compared to direct classification heads. \cref{fig:fig-2-atac-net} provides the overall structure of ATAC-Net.

\subsection{Weaker-Supervision Heads}
Given an input image $x$ and a classifier $\phi$, we can generate class probabilities $P(y_i|x)$ for the classification task. We can identify the most influential visual features from the image that result in a prediction for class $y_i$ by backpropagating $\phi(x)$ w.r.t. the output confidence scores; this helps get a visual explanation behind the network's classification.

Explainable methods like Grad-Gam \cite{grad-cam}, and SHAP \cite{shap} allows for visualising the most importable regions in an input image for predictions. The visualisation of Grad-Cam maps, which can be generated using  \cref{eq1} and \cref{eq2}  for class Dog, shown in \cref{fig:fig-1-grad-cam}, by zooming in on these most active regions, we observe that confidence scores for an object's classification are higher than the raw image. The Grad-Cam output on the zoomed sample provides evidence for comparison, showing more of the object features activated compared to the original input sample image $x$.

\begin{equation}
    \label{eq1}
    \alpha_c^k = \frac{1}{Z}\sum_i\sum_j\frac{\partial y^c}{\partial A_{ij}^k}
\end{equation}
\begin{equation}
    \label{eq2}
    L_{Grad-CAM}^c = ReLU(\sum_k\alpha_k^cA^k)
\end{equation}

Considering better outputs from $\phi$ for given cropped inputs $(x_c)$, we propose an attention-guided augmentation to learn the activation map of the features contributing towards the object class (anomaly here) via self-attention \cite{self-atten-zhao}. Furthermore, augmenting a closer crop, the model reiterates over the crop to learn the target object better even with high background noise within the input; closer cropping w.r.t. the anomaly region provides better ROI analysis using the attention features. Thus, we introduce our architecture ATAC-Net (Attention-based Anomaly Cropping Network)

The proposed ATAC-Net maps the feature extraction outputs to highlight the region representing the anomaly for a given input image $x \in R^{H \times W \times C}$. We define our network $\gamma$, comprising of three components: 1) feature extraction pipeline $\phi(x;\theta_f)$, which generates latent representation for given input image $x$, attended by 2) the proposed attention-augmentation module $\sigma(\phi(x);\theta_a)$ comprising of convolution layers followed by self-attention block to generate the saliency map corresponding to presence of anomalies within $x$. The augmentation module guides the feature extraction pipeline to reiterate $\phi$ through a close crop augmentation $x_c \in R^{H \times W \times C}$  (scaled via interpolation) using $\sigma$. Finally, we have our 3) anomaly scoring mapper $\tau(\sigma(\phi(x;\theta_f);\theta_a);\theta_s)$ to map the attention features into a score map with each element in the anomaly score mapping corresponding to a larger patch in the input sample $x$. End-to-end training of this pipeline results in learning the anomaly representation better and guides the architecture to find the anomalous regions without explicitly providing any ground truth ROIs.

\begin{equation}
    \label{eq3}
    a_{mp1} = \tau(\sigma(\phi(x;\theta_f);\theta_a);\theta_s)
\end{equation}
\begin{equation}
    \label{eq4}
    C_{map}' = \frac{1}{C}\sum_{i=1}^{C}\sigma(\phi(x; \theta_f);\theta_a)
\end{equation}
\begin{equation}
    \label{eq4.1}
    C_{map} = max_{dim=1}(C_{map}') > \omega
\end{equation}
\begin{equation}
    \label{eq5}
    x_c = crop(x; (\arg\min_{x,y}(C_{map}), \arg\max_{x,y}(C_{map})))
\end{equation}
\begin{equation}
    \label{eq6}
    a_{mp2} = \tau(\sigma(\phi(x_c;\theta_f);\theta_a);\theta_s)
\end{equation}

In the above equation, $a_{mp1}$ and $a_{mp2}$ represent the anomaly score maps, with each score on the spatial map corresponding to a patch on the input samples $x$ and crop $x_c$, respectively. \cref{eq5} describes how we obtain the cropped input $x_c$ using the channel-wise mean of the attention features from $\sigma$, which is then normalized to $0-1$; further, a threshold parameter ($\omega$) on the normalized map is used to improve the localization which helps determine "active" regions attended by $\sigma$ as shown in \cref{eq4}, \cref{eq4.1}. To obtain the final anomaly score value as a scalar, we use the $top-K$ approach over both anomaly maps to take the mean of the most significant $K$ values as the anomaly score. This is done to tackle the inconsistency of abnormalities found in various samples. Some samples may have extensive areas of irregularity, while others may only have small ones. Due to this, considering a mean or sum over the whole score map would greatly hinder the results in finding an optimal threshold point to differentiate based on these scores. The final $top-K$ mean of $a_{mp1}$ and $a_{mp2}$ produces two scalars whose mean is the final anomaly score produced by ATAC-Net. The same can be understood using \cref{eq7}, \cref{eq8}, and \cref{eq9} as follows:

\begin{equation}
    \label{eq7}
    T_{K\{i\}} = \{x_i \in a_{mp\{i\}} | i \in \arg max_{j=1}^Kx_j\}
\end{equation}
\begin{equation}
    \label{eq8}
    a_{mp\{i\}}^K = \frac{1}{K}\sum_{j=1}^KT_{K\{i\}j}
\end{equation}
\begin{equation}
    \label{eq9}
    \gamma(x;\theta) = \frac{1}{2}(a_{mp1}^K + a_{mp2}^K)
\end{equation}

Using the anomaly score generated via $\gamma$, we can identify image anomaly via the magnitude of the score. This scoring is trained through Deviation loss (discussed ahead). Most techniques rely on backpropagation to find the region where the anomaly lies and provide for interpretability, in contrast ATAC-Net learns to find the anomaly regions within input while learning the attention-augmentation module.

\subsection{Deviation Loss}
\begin{figure}[t]
    \centering
    \includegraphics[width=0.71\linewidth]{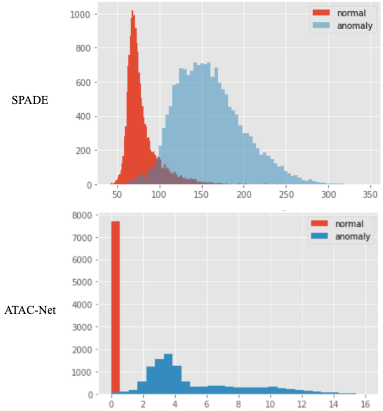} \rule{0.1\linewidth}{0pt}
    \caption{Histogram of anomalous and normal samples on a test dataset between SPADE and ATAC-Net}
    \label{fig:fig3-spade-atac}
\end{figure}

With the generated anomaly scores from the network, we require some guidance via a loss function; supervised algorithms rely on assuming the "anomaly" and "normal" samples as categories. The distinction is not well drawn for the high imbalance cases when we train the model for weak supervision and thus require excessive hyperparameter tuning. Unsupervised techniques generally try to get a distance-based measure where the idea is to reduce the distances between a set of "normal" samples during training. Once any anomalous sample comes up, a deviation is expected, and based on an optimal threshold, we classify the anomaly. Even though the scoring strategy works well, threshold selection is complex at times \cref{fig:fig3-spade-atac}. To solve this, we inculcate the deviation loss \cite{pang-devloss} where the key objective is to assume that all "normal" samples lie in a standard Gaussian distribution such that their generated anomaly scores have zero mean ($\mu = 0$) and unit standard deviation ($\sigma = 1$).  

The idea is to use the deviation of the new incoming samples \cref{eq10}, using the previously defined distribution for "normal" distribution samples. This way, we give a higher value for the deviated samples in a contrastive loss setting, i.e., anomalous samples produce much higher deviation scores than "normal" samples. The final loss can be summarized using \cref{eq11} displayed ahead.

\begin{equation}
\label{eq10}
    dev(x) = \frac{\gamma(x;\theta) - \mu_R}{\sigma_R}
\end{equation}
\begin{equation}
    \small
    \label{eq11}
    \begin{split}
    L(x_i, \mu_R, \sigma_R, \theta) = (1-y_i)|dev(x)|  \\
        + y_i(k-|dev(x)|)
    \end{split}
\end{equation}


For the defined loss in \cref{eq11}, if $ y_i  = 0$, the loss would push closer to zero, thus making the "normal" sample prediction the same. If $y_i  = 1$, the loss will move the tampered sample's anomaly score towards the defined cut-off constant $k$. Allowing for the model to learn the anomaly scoring while maintaining the gap between the "normal" and "tampered" samples, \cref{fig:fig3-spade-atac} illustrates the same observation over a private document-images dataset consisting of 7.7K genuine and 11.5K anomalies. To demonstrate the effectiveness of weak supervision with deviation loss compared to other cluster-based distance losses \cite{padim}, we have included the anomaly score variation when using an unsupervised algorithm, showing how much more distinguishable the results are visually (discussed ahead in results).

\section{Experimentation}
\label{sec:expts}
Three publicly available datasets have been tested to check the proposed methods' efficacy, each being a real-world dataset containing subtle anomaly differences helping understand the models' impact in natural settings for various tasks. For this, we consider Industrial and medical datasets to test. Since these fields cover the most impactful applications of anomaly detection and allow for agility checks of the model, this section discusses the implementation details followed for ATAC-Net, comparing networks in the weak supervision domain. We use an Nvidia RTX 2080Ti GPU with 11GB of memory for all the presented experiments in the paper.

\begin{figure}[t]
    \centering
    \includegraphics[width=0.66\linewidth]{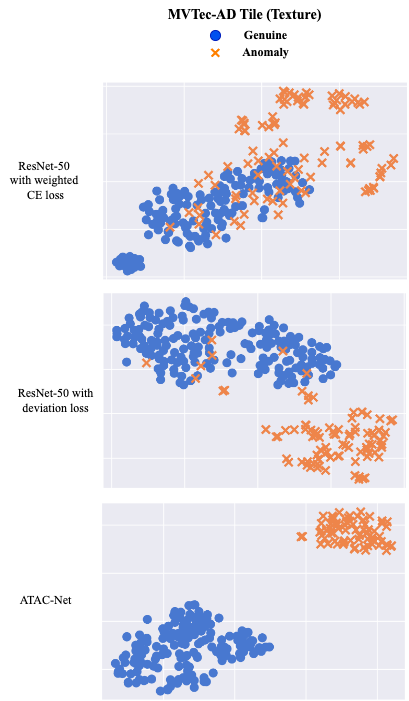} \rule{0.9\linewidth}{0pt} 
    \caption{t-SNE plots between different training versions of a baseline ResNet-50. ATAC-Net shows the effect of attention-based cropping, which helps better distinguish between anomalies and normal samples.}
    \label{fig:fig4-tsne}
\end{figure}

\subsection{Datasets}
We use the MVTec-AD \cite{mvtecad}, Head-CT \cite{datasetsdeeponeclass} and Brain-MRI \cite{datasetsdeeponeclass}. MVTec-AD dataset contains various types of surfaces and object-level anomalies, which can help determine how the model can handle subtle variations. It contains 15 different classes, with five surface-level and ten object-level. Correspondingly, the medical datasets, namely, Head-CT and Brain-MRI, each have the presence of anomalies in their respective visual modality, making them viable for detection tasks using the proposed architecture, thus helping the medical community with automated detection. The proposed network can detect deviations from new samples by incorporating both normal sample deviations and known tamper samples where ($||normal|| >> ||tamper||$). It achieves this by training on normal data to learn patch-wise score learning, like unsupervised methods and leveraging known tampers through deviation loss for detection in the datasets. To leverage the weak-supervision strategy, we leverage two different settings: 1) using only one anomaly sample compared to 2) ten anomalies to observe model performance in weak data settings, selecting the anomalies randomly in each. Further, the cut-mix \cite{cutmix} algorithm is also incorporated to add some pseudo/artificial anomalies for learning the primary cut-paste distribution over the object/surface. 

\begin{table*}
\scriptsize
    \setlength{\tabcolsep}{5pt}
    \renewcommand{\arraystretch}{1.1}
  \centering
  \begin{tabular}{c|m{4em}|cccc||c|cccc||c}
    \hline
    \multirow{2}{*}{Dataset} & \multirow{2}{4em}{\hfil No. of \\ Anomalies} & \multicolumn{5}{|c|}{One Anomaly Training} & \multicolumn{5}{|c}{Ten Anomaly Training}  \\\cline{3-12}
                             &                                              & Dev-net & DRA             & SAOE  & Cut-Mix & \textbf{ATAC-Net} & Dev-net        & DRA            & SAOE  & Cut-Mix & \textbf{ATAC-Net}   \\
    \hline 
    Carpet                   & \hfil 5                                      &  0.746  &  \textbf{0.859} & 0.766 & 0.734   & 0.849             & 0.867          & \textbf{0.940} & 0.755 & 0.803            & 0.924               \\\cline{1-2}
    Grid                     & \hfil 5                                      &  0.891  &  \textbf{0.972} & 0.921 & 0.935   & 0.918             & 0.967          & 0.987          & 0.952 & 0.931            & \textbf{0.988}      \\\cline{1-2}
    Leather                  & \hfil 5                                      &  0.873  &  0.989          & 0.993 & 0.957   & \textbf{0.994}    & 0.999          & 1.000          & 1.000 & 0.984            & \textbf{1.000}      \\\cline{1-2}
    Tile                     & \hfil 5                                      &  0.752  &  0.965          & 0.935 & 0.942   & \textbf{0.980}    & 0.987          & 0.994          & 0.944 & 0.935            & \textbf{1.000}      \\\cline{1-2}
    Wood                     & \hfil 5                                      &  0.900  &  0.985          & 0.948 & 0.893   & \textbf{0.987}    & \textbf{0.999} & 0.998          & 0.976 & 0.988            & 0.996               \\\cline{1-2}
    Bottle                   & \hfil 3                                      &  0.976  &  1.000          & 0.989 & 0.984   & \textbf{1.000}    & 0.993          & 1.000          & 0.998 & 0.991            & \textbf{1.000}      \\\cline{1-2}
    Capsule                  & \hfil 5                                      &  0.564  &  0.631          & 0.611 & 0.582   & \textbf{0.735}    & 0.865          & \textbf{0.935} & 0.850 & 0.914            & 0.934               \\\cline{1-2}
    Pill                     & \hfil 7                                      &  0.769  &  \textbf{0.832} & 0.652 & 0.649   & 0.821             & 0.866          & 0.904          & 0.872 & 0.852            & \textbf{0.921}      \\\cline{1-2}
    Transistor               & \hfil 4                                      &  0.722  &  0.668          & 0.680 & 0.715   & \textbf{0.787}    & 0.924          & 0.915          & 0.860 & 0.903            & \textbf{0.969}      \\\cline{1-2}
    Zipper                   & \hfil 7                                      &  0.922  &  0.984          & 0.970 & 0.909   & \textbf{0.988}    & 0.990          & 1.000          & 0.995 & 0.989            & \textbf{1.000}      \\\cline{1-2}
    Cable                    & \hfil 8                                      &  0.783  &  0.876          & 0.819 & 0.856   & \textbf{0.904}    & 0.892          & 0.909          & 0.862 & 0.864            & \textbf{0.983}      \\\cline{1-2}
    Hazelnut                 & \hfil 4                                      &  0.979  &  0.977          & 0.961 & 0.973   & \textbf{0.993}    & 1.000          & 1.000          & 1.000 & 1.000            & \textbf{1.000}      \\\cline{1-2}
    Metal nut                & \hfil 4                                      &  0.876  &  0.948          & 0.922 & 0.853   & \textbf{0.963}    & 0.991          & 0.997          & 0.976 & 0.952            & \textbf{1.000}      \\\cline{1-2}
    Screw                    & \hfil 5                                      &  0.399  &  \textbf{0.903} & 0.653 & 0.610   & 0.734             & 0.970          & 0.977          & 0.975 & 0.966            & \textbf{0.997}      \\\cline{1-2}
    Toothbrush               & \hfil 1                                      &  0.753  &  0.650          & 0.686 & 0.594   & \textbf{0.864}    & 0.860          & 0.826          & 0.865 & \textbf{0.901}   & 0.879               \\\cline{1-2}
    \hline
    \textbf{MVTec-AD}                 & \hfil 15                                     &  0.794  &  0.883          & 0.834 & 0.812   & \textbf{0.901}    & 0.945          & 0.959          & 0.926 & 0.932            & \textbf{0.973}      \\\cline{1-2}
    \textbf{Brain-MRI}                & \hfil 1                                      &  0.694  &  0.744          & 0.532 & 0.631   & \textbf{0.833}    & 0.958          & 0.970          & 0.900 & 0.899            & \textbf{0.979}      \\\cline{1-2}
    \textbf{Head-CT}                  & \hfil 1                                      &  0.742  &  0.796          & 0.597 & 0.570   & \textbf{0.864}    & 0.982          & 0.972          & 0.935 & 0.913            & \textbf{0.985}      \\\cline{1-2}
    \hline
  \end{tabular}
  \caption{Comparison of ATAC-net with SOTA weak-supervision methods over three datasets, two settings of weak supervision are tested through this for robustness check. all these methods have the familiar computation, using ResNet50 as backbone}
  \label{table:atacnet-compa}
\end{table*}

\begin{figure}
    \centering
    \includegraphics[width=1.2\linewidth, angle=90, origin=c]{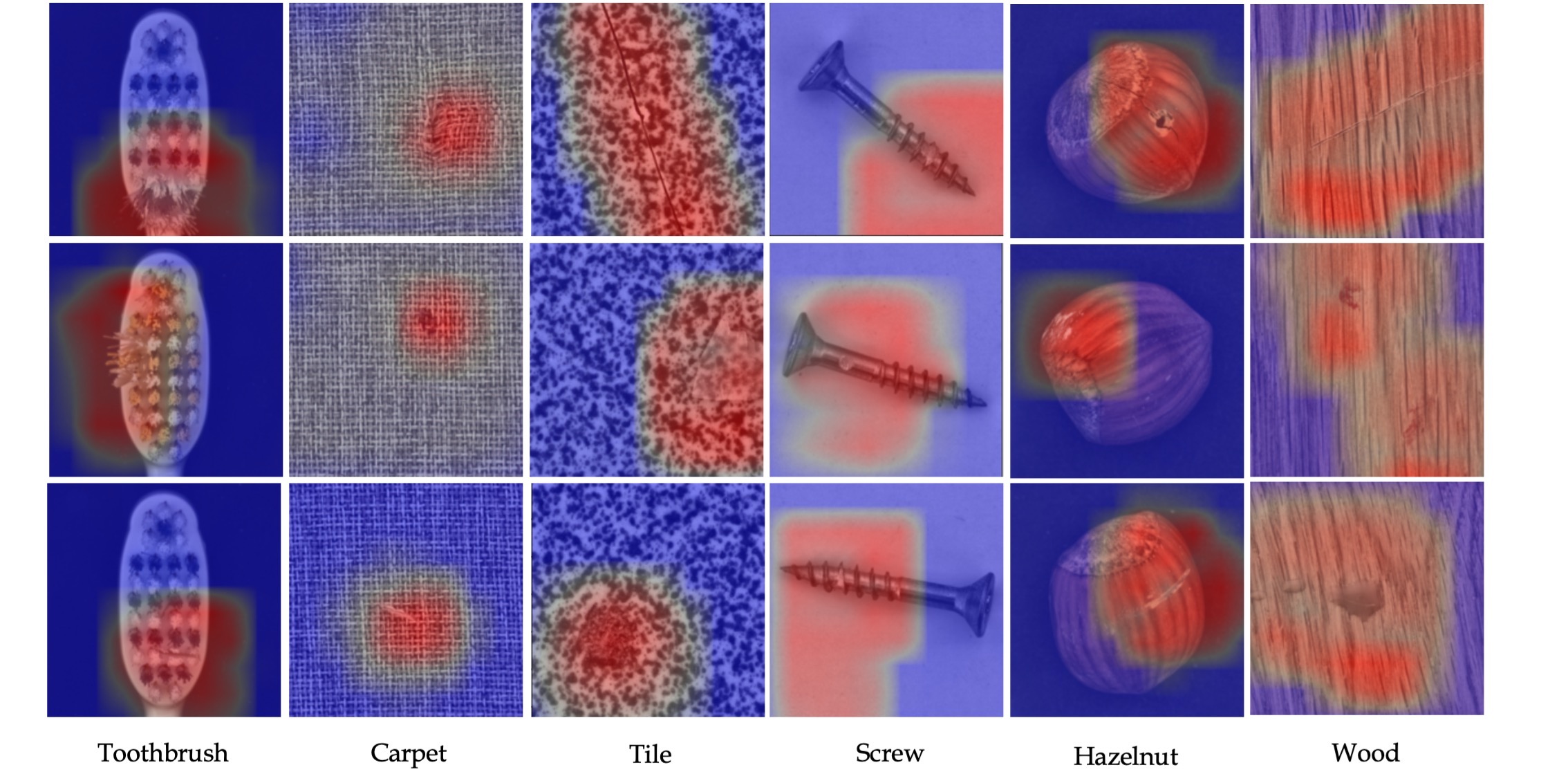}
    \caption{Anomalous regions detected by the attention-cropping mechanism from ATAC-Net. The heatmaps are binarized using a threshold after normalizing for the coordinate points extraction}
    \label{fig:atn-maps}
\end{figure}

\subsection{Training Details}
The proposed technique comprises a ResNet-50 \cite{resnet} as the backbone for our feature extraction, with $64$ channels in the convolution layers and self-attention layers added ahead of the extractor for attention cropping mechanism and zoom view generation for reiteration operation. Eq.(\ref{eq4}) sets a threshold value $\omega$ to 40\%, based on the observed loss convergence. We also apply the Cut-Mix algorithm to add pseudo anomalies during training to compensate for the data imbalance and support weak supervision. 

For comparison against other techniques which follow anomaly score-based detection, we consider the top 10\% of the scores, i.e., $top-K$ scores of the map \cite{ding-shen}\cite{pang-devloss}, to find the final anomaly score. Further, all experiments and comparisons are done at a fixed spatial resolution of $224$x$224$. We set $k$ to 10 in \cref{eq11} to differentiate the anomaly scores via deviation loss. The model is trained for $30$ epochs with a batch size of $16$ using the train set given by the respective datasets and compared with other SOTA models on the test sets. We also utilize the learning rate decay by a factor of 0.1 at a step size of 10 epochs. We use Adam \cite{adam} optimizer to update the parameters, starting at an initial learning rate $1e^{-3}$.

We compare the proposed method with all other methods using weak-supervision domains, including the likes of some SOTA techniques like DevNet \cite{pang-devloss}, DRA \cite{ding-shen}, SAOE \cite{cutpast} \cite{datasetsdeeponeclass} \cite{csi}, and a weighted categorical cross-entropy loss \cite{weighted-ce} trained Cut-Mix (using the same feature extractor ResNet-50) without using the anomaly scoring map strategy through addition of a MLP unit to act as a binary classifier. DevNet algorithm devises the deviation loss for anomaly detection for weak supervision, followed up with DRA with changes in the plain feature extraction pipeline (compared to \cite{pang-devloss}) to detect anomalies more robustly. SAOE, on the other hand, refers to using synthetic anomaly generation and outlier exposure to improve the training. Cut-Mix works by creating pseudo anomalies by selecting and pasting random input parts with other regions to simulate an anomaly. Further, we compare ATAC-Net with SOTA un-supervised techniques over the industrial dataset for surface and object-level detection accuracy in \cref{table:atacnet-typewise}. The area under the ROC curve is considered the standard metric for each table's shown comparisons.

\begin{table}[h]
    \scriptsize
    \centering
    \label{table:table-2}
    \setlength{\tabcolsep}{1.25pt}
    \renewcommand{\arraystretch}{1.20}
    \begin{tabular}{c|c|ccccccc}
    \hline
    \multicolumn{2}{c}{Category}                    & $||A||$         & Patch-SVDD     & SPADE          & PaDiM     & PatchCore     & MemSeg          & ATAC-Net       \\
    \hline
    \multirow{6}{*}{Texture}    &  Carpet             & 5               & 0.929          & 0.928          & 0.988   & 0.984     & \textbf{0.996}         & 0.924          \\\cline{2-2}
                                &  Grid               & 5               & 0.946          & 0.473          & 0.942   & 0.959     & 0.974         & \textbf{0.988} \\\cline{2-2}
                                &  Leather            & 5               & 0.909          & 0.954          & 0.996   & 1.000     & 1.000         & \textbf{1.000} \\\cline{2-2}
                                &  Tile               & 5               & 0.978          & 0.965          & 0.974   & 1.000     & 1.000         & \textbf{1.000} \\\cline{2-2}
                                &  Wood               & 5               & 0.965          & 0.958          & 0.993   & 0.991     & 0.991         & \textbf{0.996} \\\cline{2-9}
                                &  \multicolumn{2}{c}{\textbf{mean}}    & 0.945          & 0.929          & 0.959     & 0.987     & \textbf{0.992}         & 0.982 \\
    \hline
    \multirow{11}{*}{Object}    & Bottle              & 3               & 0.986          & 0.972          & 0.991   & 1.000     & 1.000         & \textbf{1.000} \\\cline{2-2}
                                & Capsule             & 5               & 0.767          & 0.897 & 0.927   & 0.982     & \textbf{0.993}         & 0.934          \\\cline{2-2} 
                                & Pill                & 7               & 0.861          & 0.801 & 0.939   & 0.920     & \textbf{0.972}         & 0.921          \\\cline{2-2} 
                                & Transistor          & 4               & 0.915          & 0.903          & 0.976 & \textbf{1.000}  & 0.986     & 0.969          \\\cline{2-2} 
                                & Zipper              & 7               & 0.979          & 0.966          & 0.882   & 0.985     & 0.994         & \textbf{1.000} \\\cline{2-2} 
                                & Cable               & 8               & 0.903          & 0.848          & 0.878   & 0.990     & 0.982         & \textbf{0.983} \\\cline{2-2} 
                                & Hazelnut            & 4               & 0.920          & 0.881          & 0.964   & 1.000     & 1.000         & \textbf{1.000} \\\cline{2-2} 
                                & Metal nut           & 4               & 0.940          & 0.710          & 0.989   & 0.994     & 1.000         & \textbf{1.000} \\\cline{2-2} 
                                & Screw               & 5               & 0.813          & 0.667          & 0.845   & 0.960     & 0.978         & \textbf{0.997} \\\cline{2-2} 
                                & Toothbrush          & 1               & \textbf{1.000} & 0.889          & 0.942   & 0.933     & 1.000         & 0.879          \\\cline{2-9} 
                                &  \multicolumn{2}{c}{\textbf{mean}}  & 0.908          & 0.976          & 0.975     & 0.976     & \textbf{0.991}         & 0.969          \\
    \hline
    \multicolumn{2}{c}{MVTec-AD}                    & 15              & 0.921          & 0.854          & 0.948     & 0.980     & \textbf{0.991}         & 0.973 \\
    \hline
    \multicolumn{3}{c}{Inference Time (ms)}                               & 480          & 339          & 319     & 225     & 32         & \textbf{29} \\
    \hline
    \end{tabular}
    \caption{Comparison of ATAC-net with other SOTA unsupervised techniques over MVTec-AD dataset. $||A||$ represents the number of different anomalies present for that category}
    \label{table:atacnet-typewise}
\end{table}

\section{Results}
\label{sec:results}

\cref{table:atacnet-compa} compares all the mentioned algorithms using weak supervision, showing the improvements of the proposed ATAC-Net. The technique comes in close with the DRA algorithm on a few points. However, overall, the zoomed view technique achieves better results and even provides interpretability by the model for detected anomalies through a saliency map. The saliency maps for a few surface and object classes of the MVTec-AD dataset are in \cref{fig:atn-maps}. To understand more about the explanability provided by the model, we additionally observe the t-SNE \cite{tsne} plot for the difference between the base ResNet-50 pipeline with weighted cross-entropy as a baseline compared with the same feature extractor using deviation loss and finally with our proposed attention-augmentation module. All these accumulated results can be observed in \cref{fig:fig4-tsne} with each method using the Cut-Mix algorithm and ten anomalous samples for weak supervision. Given the setting in all the mentioned datasets, ATAC-Net beats all the current weak supervision-based models in determining the anomalies.

Also, for the graph in \cref{fig:fig3-spade-atac}, we observed that for a sizeable internal dataset (19K samples) comprising anomalies on the document surface, ATAC-Net maintained an easy-to-determine threshold between the "normal" and tampered class by limiting the "normal" samples class in closer proximity to 0. Whereas using SPADE \cite{spade}, the results show a considerable overlap of "normal" and "tampered" sample scores, complicating the process of finding the optimal threshold every time.

Comparing the weak-supervised ATAC-Net with some SOTA un-supervised models like Patch-SVDD \cite{svdd}, SPADE \cite{spade}, PaDiM \cite{padim}, PatchCore \cite{total_recall}, and MemSeg \cite{memseg} (using ResNet-50), in \cref{table:atacnet-typewise}, we provide a comparison between the object-level and the surface-level anomaly detection scoring. From the results, it is viable that ATAC-Net is very comparable on both surface level and object class, with many classes surpassing the unsupervised methods by getting a 100\% AUROC score, failing significantly only on the toothbrush object. This case might be due to the cut-mix augmentation providing an out-of-context pasting for this object. Overall, the results are viable and provide better differentiability from other stated methods (\cref{fig:fig3-spade-atac}, \cref{fig:fig4-tsne}). Further, the noise-based foreground anomalies \cite{memseg} can act as a better augmentation along with Cut-Mix for training as done by MemSeg \cite{memseg}.

Despite the accurate localization of anomalous regions, it is essential to understand that the augmented module is localizing on a large scale, giving out a little closer view; this gates us from generating pixel-level AUROC scores by just using the anomaly segments from attention maps. However, observing \cref{fig:atn-maps}, the self-attention module, when left with the provided contrastive setting, can reiterate over the generated saliency map at each step to learn the position of anomaly pretty accurately.

\section{Conclusion And Future Works}
\label{sec:conc}
Throughout this paper, we discussed how a zoomed-in view of input allows for better anomaly detection compared to standard feature extraction pipelines when given to a network. The need for anomaly supervision is ignored at many points by the advent of unsupervised algorithms, where they work simply by observing only the "normal" samples of a distribution; ATAC-Net showed, along with some other weakly supervised techniques, that getting a very few samples in training can provide accurate inferencing with easy determination in a real-world setting. Further, the lack of proper boundaries in anomalous and normal samples is handled here via deviation loss training. We also outperform some SOTA models' results over three real-life datasets. Despite these results, further work is still required to localise the anomalies better. For instance, it is possible to train the attention-augmentation pipeline with the ground-truth anomaly segmentation maps to estimate the locals better. Further, it is possible to test the unsupervised techniques by generating anomaly maps and again form the close look augmentation to better attend to the anomaly scoring and distinguish the embedding space accordingly.  

Supplementary material to dive in further is available at this link on \href{https://sigport.org/documents/atac-net-zoomed-view-works-better-anomaly-detection}{sigport}
{\small
\bibliographystyle{ieeetr}
\bibliography{main.bib}

\begin{thebibliography}{10}

\bibitem{grad-cam}
R.~R. Selvaraju, M.~Cogswell, A.~Das, R.~Vedantam, D.~Parikh, and D.~Batra, ``Grad-cam: Visual explanations from deep networks via gradient-based localization,'' in {\em 2017 IEEE International Conference on Computer Vision (ICCV)}, pp.~618--626, 2017.

\bibitem{total_recall}
K.~Roth, L.~Pemula, J.~Zepeda, B.~Schölkopf, T.~Brox, and P.~Gehler, ``Towards total recall in industrial anomaly detection,'' in {\em 2022 IEEE/CVF Conference on Computer Vision and Pattern Recognition (CVPR)}, pp.~14298--14308, 2022.

\bibitem{fflow}
J.~Yu, Y.~Zheng, X.~Wang, W.~Li, Y.~Wu, R.~Zhao, and L.~Wu, ``Fastflow: Unsupervised anomaly detection and localization via 2d normalizing flows,'' {\em CoRR}, vol.~abs/2111.07677, 2021.

\bibitem{padim}
T.~Defard, A.~Setkov, A.~Loesch, and R.~Audigier, ``Padim: A patch distribution modeling framework for anomaly detection and localization,'' in {\em Pattern Recognition. ICPR International Workshops and Challenges} (A.~Del~Bimbo, R.~Cucchiara, S.~Sclaroff, G.~M. Farinella, T.~Mei, M.~Bertini, H.~J. Escalante, and R.~Vezzani, eds.), (Cham), pp.~475--489, Springer International Publishing, 2021.

\bibitem{mahab-dist}
H.~Ghorbani, ``Mahalanobis distance and its application for detecting multivariate outliers,'' {\em Facta Universitatis Series Mathematics and Informatics}, vol.~34, p.~583, 10 2019.

\bibitem{resnet}
K.~He, X.~Zhang, S.~Ren, and J.~Sun, ``Deep residual learning for image recognition,'' in {\em 2016 IEEE Conference on Computer Vision and Pattern Recognition (CVPR)}, pp.~770--778, 2016.

\bibitem{memseg}
M.~Yang, P.~Wu, and H.~Feng, ``Memseg: A semi-supervised method for image surface defect detection using differences and commonalities,'' {\em Engineering Applications of Artificial Intelligence}, vol.~119, p.~105835, 2023.

\bibitem{norm-flows}
D.~J. Rezende and S.~Mohamed, ``Variational inference with normalizing flows,'' in {\em Proceedings of the 32nd International Conference on International Conference on Machine Learning - Volume 37}, ICML'15, p.~1530–1538, JMLR.org, 2015.

\bibitem{cflowad}
D.~Gudovskiy, S.~Ishizaka, and K.~Kozuka, ``Cflow-ad: Real-time unsupervised anomaly detection with localization via conditional normalizing flows,'' in {\em 2022 IEEE/CVF Winter Conference on Applications of Computer Vision (WACV)}, (Los Alamitos, CA, USA), pp.~1819--1828, IEEE Computer Society, jan 2022.

\bibitem{ding-shen}
C.~Ding, G.~Pang, and C.~Shen, ``Catching both gray and black swans: Open-set supervised anomaly detection,'' in {\em 2022 IEEE/CVF Conference on Computer Vision and Pattern Recognition (CVPR)}, (Los Alamitos, CA, USA), pp.~7378--7388, IEEE Computer Society, jun 2022.

\bibitem{pang-devloss}
G.~Pang, C.~Ding, C.~Shen, and A.~van~den Hengel, ``Explainable deep few-shot anomaly detection with deviation networks,'' {\em CoRR}, vol.~abs/2108.00462, 2021.

\bibitem{focal-loss}
T.-Y. Lin, P.~Goyal, R.~Girshick, K.~He, and P.~Dollár, ``Focal loss for dense object detection,'' {\em IEEE Transactions on Pattern Analysis and Machine Intelligence}, vol.~42, no.~2, pp.~318--327, 2020.

\bibitem{zhao-l1-loss}
H.~Zhao, O.~Gallo, I.~Frosio, and J.~Kautz, ``Loss functions for neural networks for image processing,'' 2018.

\bibitem{Zhang2022PrototypicalRN}
H.~M. Zhang, Z.~Wu, Z.~Wang, Z.~Chen, and Y.~Jiang, ``Prototypical residual networks for anomaly detection and localization,'' {\em ArXiv}, vol.~abs/2212.02031, 2022.

\bibitem{shap}
S.~M. Lundberg and S.-I. Lee, ``A unified approach to interpreting model predictions,'' in {\em Advances in Neural Information Processing Systems 30} (I.~Guyon, U.~V. Luxburg, S.~Bengio, H.~Wallach, R.~Fergus, S.~Vishwanathan, and R.~Garnett, eds.), pp.~4765--4774, Curran Associates, Inc., 2017.

\bibitem{self-atten-zhao}
H.~Zhao, J.~Jia, and V.~Koltun, ``Exploring self-attention for image recognition,'' in {\em 2020 IEEE/CVF Conference on Computer Vision and Pattern Recognition (CVPR)}, pp.~10073--10082, 2020.

\bibitem{mvtecad}
P.~Bergmann, M.~Fauser, D.~Sattlegger, and C.~Steger, ``Mvtec ad — a comprehensive real-world dataset for unsupervised anomaly detection,'' in {\em 2019 IEEE/CVF Conference on Computer Vision and Pattern Recognition (CVPR)}, pp.~9584--9592, 2019.

\bibitem{datasetsdeeponeclass}
P.~Liznerski, L.~Ruff, R.~A. Vandermeulen, B.~J. Franks, M.~Kloft, and K.~R. Muller, ``Explainable deep one-class classification,'' in {\em International Conference on Learning Representations}, 2021.

\bibitem{cutmix}
S.~Yun, D.~Han, S.~Chun, S.~J. Oh, Y.~Yoo, and J.~Choe, ``Cutmix: Regularization strategy to train strong classifiers with localizable features,'' pp.~6022--6031, 10 2019.

\bibitem{adam}
D.~P. Kingma and J.~Ba, ``Adam: {A} method for stochastic optimization,'' in {\em 3rd International Conference on Learning Representations, {ICLR} 2015, San Diego, CA, USA, May 7-9, 2015, Conference Track Proceedings} (Y.~Bengio and Y.~LeCun, eds.), 2015.

\bibitem{cutpast}
C.~Li, K.~Sohn, J.~Yoon, and T.~Pfister, ``Cutpaste: Self-supervised learning for anomaly detection and localization,'' in {\em 2021 IEEE/CVF Conference on Computer Vision and Pattern Recognition (CVPR)}, (Los Alamitos, CA, USA), pp.~9659--9669, IEEE Computer Society, jun 2021.

\bibitem{csi}
J.~Tack, S.~Mo, J.~Jeong, and J.~Shin, ``{CSI:} novelty detection via contrastive learning on distributionally shifted instances,'' {\em CoRR}, vol.~abs/2007.08176, 2020.

\bibitem{weighted-ce}
T.~H. Phan and K.~Yamamoto, ``Resolving class imbalance in object detection with weighted cross entropy losses,'' {\em CoRR}, vol.~abs/2006.01413, 2020.

\bibitem{tsne}
L.~van~der Maaten and G.~E. Hinton, ``Visualizing data using t-sne,'' {\em Journal of Machine Learning Research}, vol.~9, pp.~2579--2605, 2008.

\bibitem{spade}
N.~Cohen and Y.~Hoshen, ``Sub-image anomaly detection with deep pyramid correspondences,'' {\em CoRR}, vol.~abs/2005.02357, 2020.

\bibitem{svdd}
J.~Yi and S.~Yoon, ``Patch svdd: Patch-level svdd for anomaly detection and segmentation,'' in {\em Computer Vision -- ACCV 2020} (H.~Ishikawa, C.-L. Liu, T.~Pajdla, and J.~Shi, eds.), (Cham), pp.~375--390, Springer International Publishing, 2021.

\end{thebibliography}
}

\end{document}